# Probabilistic Logic Gate in Asynchronous Game of Life with Critical Property

Yukio-Pegio Gunji[1, *], Yoshihiko Ohzawa[1] and Terutaka Tanaka[1]

[1] Department of Intermedia, Art and Science, School of Fundamental Science and Technology, Waseda University, 3-4-1, Ohkubo, Shinjuku, Tokyo 169-8555, Japan

*Author correspondence; yukio@waseda.jp

Abstract

Metaheuristic and self-organizing criticality (SOC) could contribute to robust computation under perturbed environments. Implementing a logic gate in a computing system in a critical state is one of the intriguing ways to study the role of metaheuristics and SOCs. Here, we study the behavior of cellular automaton, game of life (GL), in asynchronous updating and implement probabilistic logic gates by using asynchronous GL. We find that asynchronous GL shows a phase transition, that the density of the state of 1 decays with the power law at the critical point, and that systems at the critical point have the most computability in asynchronous GL. We implement AND and OR gates in asynchronous GL with criticality, which shows good performance. Since tuning perturbations play an essential role in operating logic gates, our study reveals the interference between manipulation and perturbation in probabilistic logic gates.

1. Introduction

   As we confront a large amount of data, it is difficult to obtain a complete dataset, resulting in an incomplete and ill-defined problem. Therefore, metaheuristics are investigated more intensely than optimization (Erskine & Hermann, 2015). The first direction of metaheuristics is bioinspired computing based on swarm intelligence, such as the krill herd algorithm (Wang et al., 2014), monarch butterfly optimization (Chakrabarty et al., 2014), ant colony algorithm (Dorigo & Stützle, 2004), and others (Wang, Deb & Coelho, 2018). These algorithms can search for quasi-optimistic solutions by balancing the global goal of a group with internal perturbations. The second direction is the development of biological computing, in which computing is implemented by living biological and/or chemical material, such as BZ pattern computing (Toth et al. 2009), physarum computing (Nakagaki, Yamada & Toth, 2000; Takamatsu et al., 2001; Tero et al., 2010; Tsuda, Aono & Gunji, 2004; Gunji et al., 2008; 2011; Adamatzky, 2010; Aono et al., 2015; Zhu et al., 2018), fungal computing (Adamatzky, 2018; Adamatzky et al., 2020) and soldier crab computing (Gunji, Nishiyama & Adamatzky, 2011; Nishiyama, Gunji & Adamatzky, 2012). Since a living system maintains a wholeness as a living unity and confronts perturbed surroundings, it is obliged to balance the global goal as a system with surrounding perturbations (Cordero, 2017; Erskine & Hermann, 2015; Gunji et al., 2020b). Thus, living computing is expected to solve an ill-defined problem and to obtain a quasi-optimal solution. Since dissipative structures such as the BZ pattern are surrounded by a specific boundary condition, they also contain a wholeness to some extent. In that sense, they are similar to quasi-biological computing.

   The third direction is the reevaluation of self-organizing criticality (SOC), in which a critical state characterized by a power law distribution is automatically obtained by balancing the global optimum with perturbation (Bak, Tang & Wiesnfeld 1987; Bak & Tang, 1989; Kauffman & Johnsen, 1991; Bak & Sneppen, 1993). Although there is a disadvantage for SOC that requires a fitness function (or potential function) that is difficult to find, the idea of criticality could contribute to the attribute of a quasi-optimal solution under perturbed conditions. Since it has been found that biological systems usually reveal the features of critical phenomena, they are expected to be systems showing SOC (e.g., Levi walk; Reynolds, 2018; Viswanathan et al. 1999; Bartumeus, 2007; Sims et al., 2008; Humphries et al., 2010; Reynolds, Schultheiss & Cheng, 2013). It leads to the idea that the results of bioinspired computing are destined to be quasi-optimal solutions if they show critical phenomena characterized by the power law. With respect to the power law distribution, the first, second and third directions of metaheuristics can be combined. This leads to the powerful idea that instead of implementing SOCs by using a specific fitness, bioinspired computing can play a more important role in generating criticality, which can reveal quasi-optimal solutions.

Quasi-optimal solutions for an ill-defined problem could be generalized by universal and efficient computational ability under perturbed natural conditions (Gunji & Uragami, 2020). For this reason, universal and efficient computing devices can be made of biological materials, such as physarum gates, solider crab gates and some physarum computers. The next question arises (Conrad, 1983; 1985): what is the essential property in bioinspired or biological computing? Bioinspired computing or biological computing is expressed as a multi-agent system whether it is a multicellular or unicellular organism under indefinite conditions (Erskine & Hermann, 2015; Cordero, 2017; Erskine & Hermann, 2015; Gunji et al., 2020b). Therefore, these systems are obliged to be probabilistic and/or asynchronous in updating due to indefinite and perturbed environments. Conversely, bioinspired and biological computing is robust due to the probabilistic and/or asynchronous property. Thus, it is interesting to study the relationship among quasi-optimal solutions, critical properties, and probabilistic and/or asynchronous updating in a computational system (Gunji, 1990; Gunji 2014a, b; Gunji & Uragami, 2020). This leads to the possibility of probabilistic logic gates implemented by asynchronously updating computations. It is important to determine the core in bioinspired computing that is robust under perturbed conditions.

To answer those questions, we take a special cellular automaton called game of life (GL) (Gardner, 1970; Berlekamp, Conway & Guy, 1982; Nordfalk & Alstrøm, 1996; Rennard, 2002). Originally, game of life was updated in a synchronous fashion and could be used as a logic gate by using a specific pattern called gliders and glider guns (Berlekamp, Conway & Guy, 1982; Rennard, 2002). Since these patterns are generated in a synchronous fashion, their behaviors can be strictly controlled by initial and boundary conditions. Although the trajectories of gliders going straight can be predicted, they can be easily broken by small perturbations. The question of how we could construct robust logic gates made of gliders arises. If possible, one can implement robust probabilistic logic gates rather than unstable deterministic logic gates. Thus, we aim to determine the core in bioinspired computing that is robust under perturbed conditions

## 2. Asynchronous Game of Life and its Logic gates

### 2-1 Phase transition and criticality

Game of life (GL) is a two-dimensional cellular automaton in which the state of a cell is either 0 or

1 (Gardner, 1970; Berlekamp, Conway & Guy, 1982; Nordfalk & Alstrøm, 1996; Rennard, 2002). The state of a cell at the $(i, j)$ site at the $t$th step is represented by $a_{i,j}^t$. To compare asynchronous GL with synchronous GL, we introduce the virtual state of a cell at the $(i, j)$ site at the $t$th step, which is represented by $b_{i,j}^t$. The transition rule of the state of a cell is defined as follows.

In the case of $a_{i,j}^t = 0$:
$$\text{If } S_{i,j}^t = 3, \text{ then } b_{i,j}^{t+1} = 1; \tag{1}$$
$$\text{Otherwise } b_{i,j}^{t+1} = 0,$$

In the case of $a_{i,j}^t = 1$:
$$\text{If } S_{i,j}^t = 5 \text{ or } 6, \text{ then } b_{i,j}^{t+1} = 1; \tag{2}$$
$$\text{Otherwise } b_{i,j}^{t+1} = 0,$$

where

$$S_{i,j}^t = \sum_{m=-1,n=-1}^{m=+1,n=+1} a_{i+m,j+n}^t - a_{i,j}^t. \tag{3}$$

The transition of synchronous GL is defined by

$$a_{i,j}^{t+1} = b_{i,j}^{t+1}. \tag{4}$$

In contrast, the transition of asynchronous GL is defined by using probability $p$ with $0 \leq p \leq 1$ (e.g., Fatès 2001; 2006; 2014) such that

$$a_{i,j}^{t+1} = a_{i,j}^t \text{ with } p; \tag{5}$$
$$a_{i,j}^{t+1} = b_{i,j}^{t+1} \text{ with } 1\text{-}p.$$

It is easy to see that there are both active and inactive (stationary) phases in asynchronous GL and that a cell in the stationary phase is not obliged to follow transition rules (1)-(3). If $p=0$, then asynchronous GL coincides with synchronous GL. If $p=1$, then all cells are always at stationary phase and preserve given initial states over time.

Figure 1 shows the snapshots of the pattern generated by asynchronous GL controlled by the probability $p$ in equation (5), where the system size is 50 by 50. It is easy to see that under small $p$, the states converge to frozen states that oscillate with period 2 at most, while under large $p$, the states chaotically change. One can see the phase transition of the generated pattern with respect to $p$.

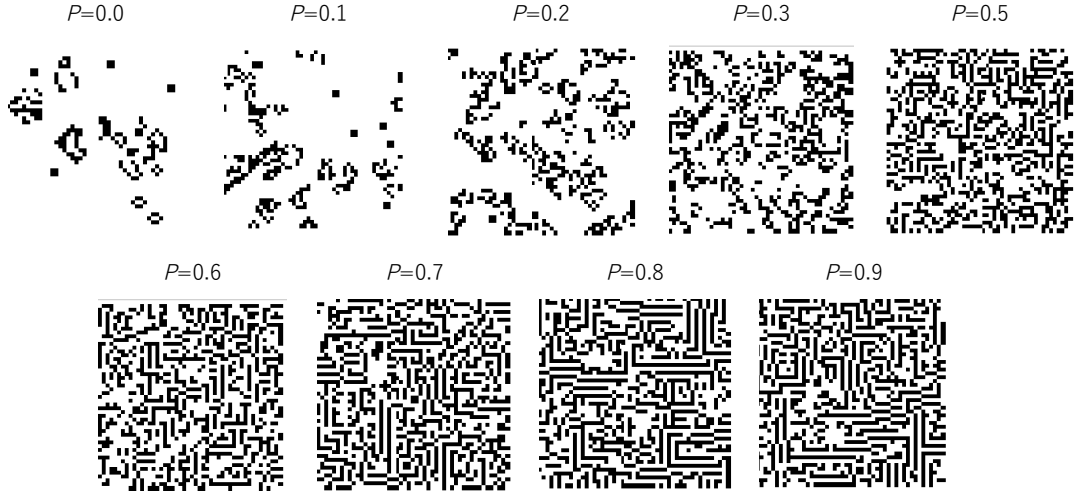

Figure 1. Snapshots of asynchronous GL controlled by the probability $p$. States 1 and 0 are represented by black squares or blanks.

Figure 2 shows the phase transition from ordered patterns to chaotic patterns with respect to the probability of the stationary phase. Patterns are generated in 500 by 500 squared cells with periodic boundaries. It is determined whether all cells converge to oscillatory states with a period equal to or smaller than 2 within 10000 steps in each trial. If true, it is counted as a frozen state. In 100 trials, the normalized number of frozen states is interpreted as the probability of the frozen state. Each circle represents the probability of the frozen state plotted against the probability of the stationary phase, $p$. To estimate the position of the critical state, the range between 0.095 and 0.155 was intensively investigated. The red curve represents a sigmoidal curve fitted to data, such as

$$y = \frac{1}{1+e^{a(x-b)}} \qquad (6)$$

where $a = 115.038$ and $b = 0.1269$. The critical value is approximately 0.13. Once the probability of the frozen state drops to 0, it is not recovered. If $p$ is larger than 0.2, the states of asynchronous GL are perpetually changed in time and space.

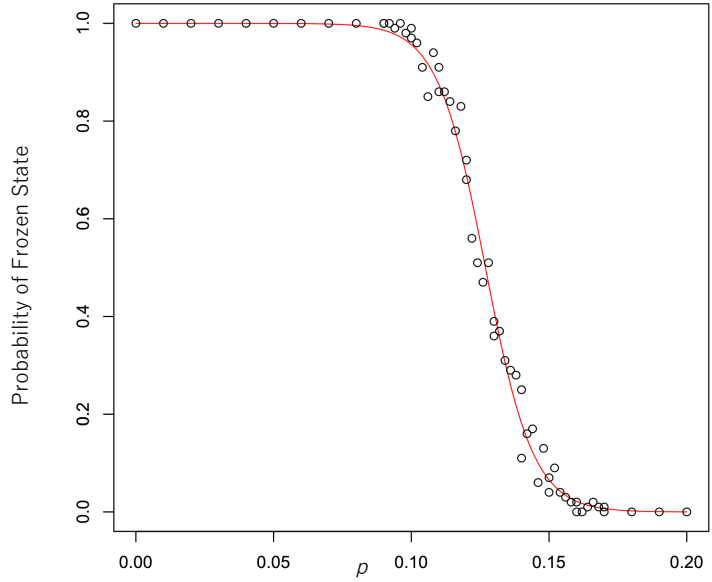

Figure 2. The probability of the frozen state plotted against *p* (the probability of the stationary phase). The red curve represents the approximation by the sigmoidal curve.

The next question arises, does the middle point between two phases, order (oscillation) and chaos, indicate the critical state. Asynchronous cellular automata showing cluster patterns or class IV patterns frequently reveal a power law distribution with a density of 1 plotted against time. Additionally, in asynchronous GL, the density against time is evaluated.

Figure 3 shows the density of the state of 1 against time in a log-log plot. The initial condition is set for the density of states of 1 at 0.590. The system size is 150 by 150 cells and *p*=0.13. The density against time is averaged over 20 trials and is represented by the green curve in Figure 3. The data are approximated by a power-law function such as

$$y = a(x - b)^c \tag{7}$$

where *a* = 0.020 and *b* = 0.081, and the exponent of the power is -0.1595.

The exponent -0.1595 could imply special meaning since this value can be found in various contexts (Domany & Kinzel, 1984; Hinrichsen, 2000; Fatès & Morvan, 2005). The model of directed percolation is defined by a squared two-dimensional system consisting of *N* by *N* cells. The state of a cell is either 0 or 1, where 1 means rocky material and 0 means a pore. The number of pores divided by $N^2$ represents the porosity. Initially, water is given in cells at the top layer. Vertically water falls through pores. If the water reaches at least one cell at the bottom layer, percolation succeeds. The normalized success rate of percolation is obtained for multiple trials, where the distribution of pores is randomly distributed for each trial. Directed percolation shows the phase transition between 100% success and 0% success of percolation. It is numerically obtained that the critical value of the porosity is 0.705 and that the density of water at the critical point is decayed in a power law fashion whose exponent is -0.1595.

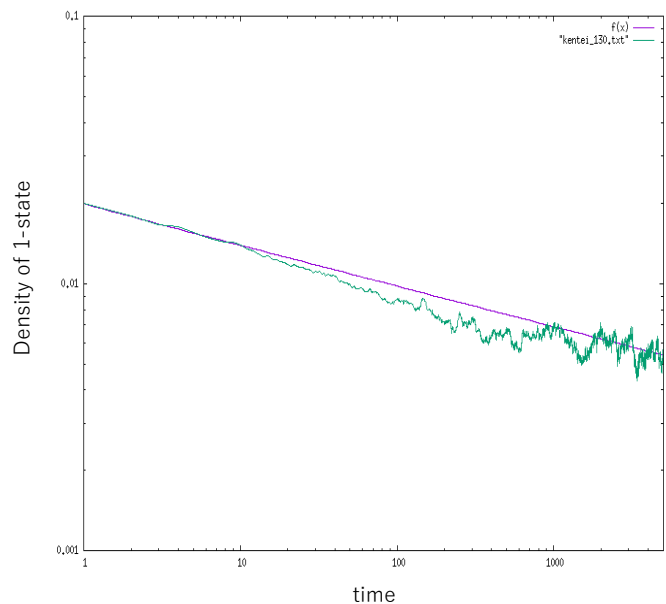

Figure 3. The density of 1 plotted against time in a log-log scale. Data averaged over 20 trials are represented by green curves, and the approximated line (power law) is represented by a purple line.

Some rules of elementary cellular automata updated asynchronously also show the phase transition with respect to the probability of stationary phase (Fatès & Morvan, 2005). At the critical point, the density of the state of 1 decays in a power law fashion whose exponent is -0.1595. If asynchronous updating is defined by the order of updating, where a map of the order of cells to the order of updating is a bijection, and if the order of updating can influence a transition rule, many elementary cellular automata show cluster-like or class IV-like patterns mixing the localized periodic patterns with chaotic patterns. Most of them show the power law decay of the density of the state of 1 with an exponent of -0.1595 (Gunji 2014a, b). Therefore, this exponent, -0.1595, might show universal properties at the critical point in the phase transition.

2-2 Computation by asynchronous GL

Synchronous GL can be used as a computing resource since mesoscopic patterns called gliders can be used as a transmission of the value. As shown in the left diagram of Figure 4, if a special configuration called a glider gun is prepared, a glider that goes straight while vibrating is perpetually generated. Indeed, if a glider is controlled, whether it is active (i.e., gliders can be generated) or inactive (i.e., gliders cannot be generated), one can implement the AND gate and NOT gate.

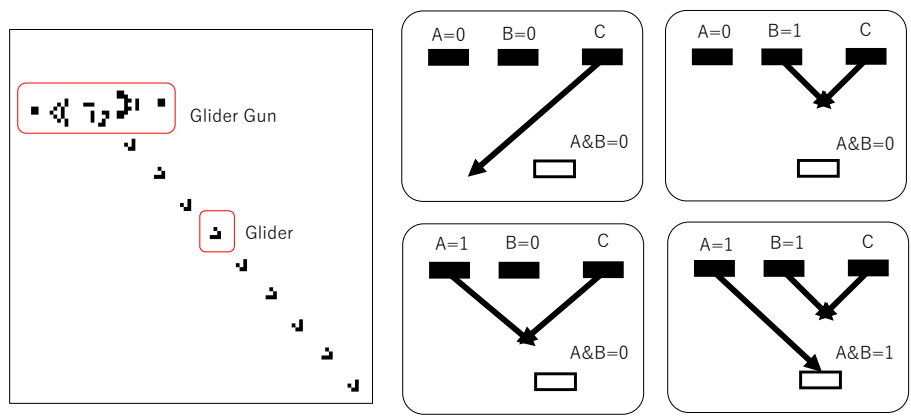

Figure 4. Glider and AND gate made of gliders in synchronous GL. For input gliders A and B set at the solid squares, output A AND B (represented by A&B) is set at the white square.

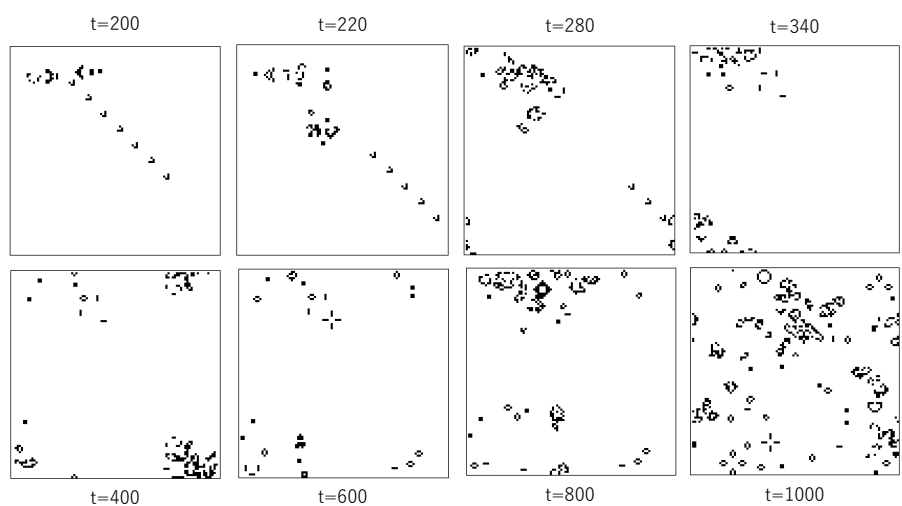

Figure 5. Gliders under the perturbation of the inversion of 0 and 1, with a probability of 0.0001.

The right diagram of Figure 4 shows the implementation of the AND gate by gliders. A dummy glider gun represented by C, hidden in the gate device, can constantly generate a glider and is prepared at the location in which input gliders from A and B can be collided by the dummy glider from C. If two gliders collide with each other, the gliders disappear. If firing a glider represents a value of 1, and no glider represents a value of 0, one can prepare the A AND B gate, as shown in the right diagram of Figure 4. This implementation reveals that if the input glider guns and the location of the output are stably controlled, synchronous GL has the computation ability. However, the gliders are unstable against perturbation.

Figure 5 shows the behavior of gliders under the perturbation when the state of a cell is inverted with a probability of 0.0001 (i.e., 0 is replaced by 1 and vice versa), where there is no perturbation before $t = 200$. Once a perturbation occurs, the glider never goes straight and is collapsed anywhere. Instead of normal gliders, broken gliders are generated and propagate radially.

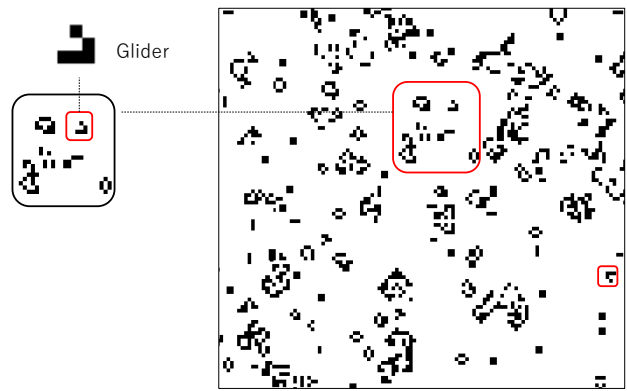

Figure 6. Rare gliders from the initial condition where 0 and 1 are randomly distributed. Gliders are surrounded by red squares.

Even if there is no perturbation, gliders are not stable from the initial condition, where 0 and 1 are randomly distributed. Gliders can be generated very rarely from the random initial condition, and they are collapsed by fragile broken gliders propagating irregularly. Figure 6 shows the general case soon after the random initial condition, in which only two gliders are generated. After this situation, local configurations fall into the oscillatory state called flickers. Under no perturbation, only strict and rigid control of gliders can implement logic gates in synchronous GL. In other words, strict and rigid computations such as synchronous GL cannot contribute to real-world computation, which always confronts perturbed conditions. Although theoretical computer scientists focus on the implementation of universal Turing machines or complete machines, one should think about robust but not highly efficient or universal computation.

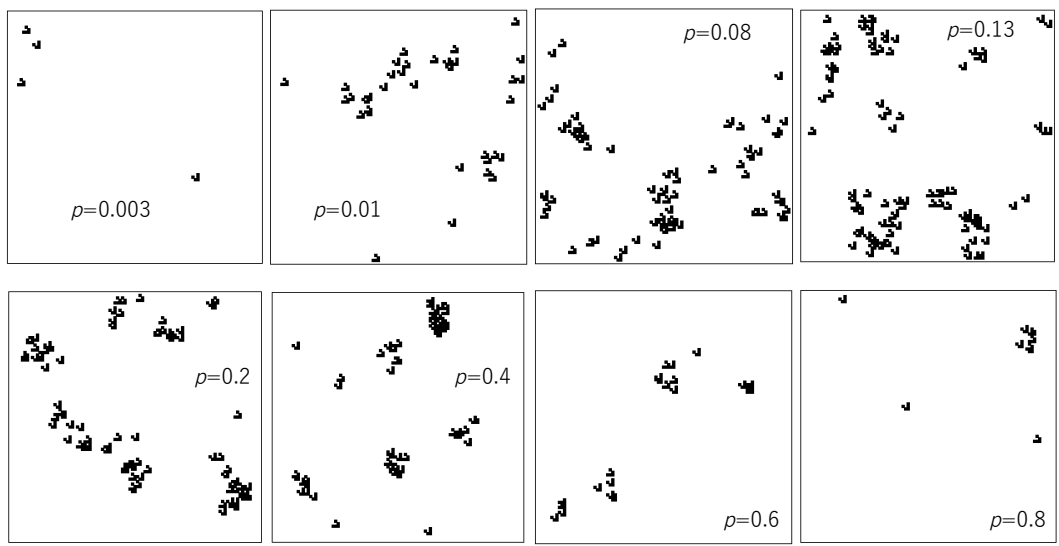

Figure 7. Occurrence of gliders in asynchronous GL. In each diagram, gliders within 100 steps from the initial condition overlap.

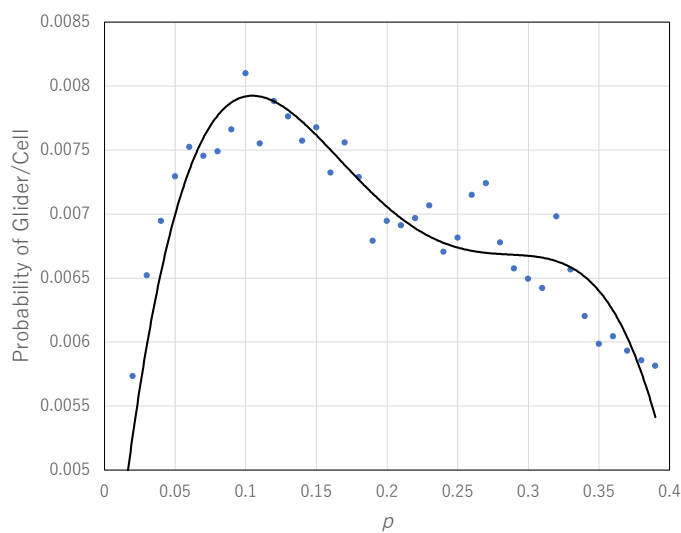

Figure 8. Probability of glider per cell plotted against the probability of the stationary phase. The peak is found close to the critical point.

The next question arises: how can probabilistic computation be implemented against perturbed conditions? Here, we implement a probabilistic logic gate made of broken gliders generated in asynchronous GL. First, we estimate how many gliders are generated in the range of phase transitions of asynchronous GL. Figure 7 shows the relationship between the occurrence of gliders and the probability of the stationary phase in asynchronous GL.

It is easy to see that the occurrence of gliders is increased in the range surrounding the critical value, $p\sim0.13$. Far from the critical value, few gliders are generated. This implies that an asynchronous GL with critical properties can be used as a probabilistic logic gate since many gliders that break and can be irregularly propagated are generated around the critical value. Figure 8 shows the normalized occurrence of gliders plotted against the probability of the stationary phase. The number of gliders per time step is normalized by the system size, which is regarded as the probability of gliders per cells. The curve represents the approximation by polynomial functions such as $y = -3.1924x^2 + 2.8931x^3 - 0.9137x^2 + 0.1109x + 0.0034$.

2-3 Logic gate in asynchronous GL

The probabilistic logic gate in asynchronous GL is implemented as follows. The square space consisting of $N^2$ cells is divided into four areas. The first area $S(A)$ is for input value $A$, the second area $S(B)$ is for input $B$, the third area $S(C)$ is for output value, and the fourth area is the rest of the $N^2$ cells. These areas are defined by:

$$S(A) = \{(i,j) | i = 1,2, \dots, \tfrac{N}{3}; j = 1, 2, \dots, \tfrac{N}{3}\}, \tag{8}$$
$$S(B) = \{(i,j) | i = \tfrac{2N}{3}, \tfrac{2N}{3}+1, \dots, N; j = \tfrac{2N}{3}, \tfrac{2N}{3}+1, \dots, N\}, \tag{9}$$
$$S(C) = \{(i,j) | i = \tfrac{N}{3}+1, \tfrac{N}{3}+2, \dots, \tfrac{2N}{3}-1; j = \tfrac{N}{3}+1, \tfrac{N}{3}+2, \dots, \tfrac{2N}{3}-1\}. \tag{10}$$

In the AND gate, *C* is indicated by *A* AND *B* (sometimes represented by *A&B*), and in the OR gate, *C* is indicated by *A* OR *B* (sometimes represented by *A||B*). In the probabilistic logic gate, the value 1 is defined by a density of 1 that exceeds the specific probability, $P_1$. When a value of 1 is given for input *X* with *X=A, B*, each cell in $S(X)$ has the state of 1 with the probability exceeding $P_1$. A value of 0 for input *X* is defined by the situation in which all cells in $S(X)$ have a state of 0. After setting input values, asynchronous GL with the probability of stationary phase $p = 0.13$ (critical value) is adapted to all cells in *T* time steps under the condition with perturbation in which state of a cell is inverted with the probability $p_{NOISE}$. If the density of $S(C)$ exceeds $P_1$ at *T*, then output *C* is 1; otherwise, *C* is 0. Note that although the definition of input 0 is consistent with that of output 0, they are different from each other.

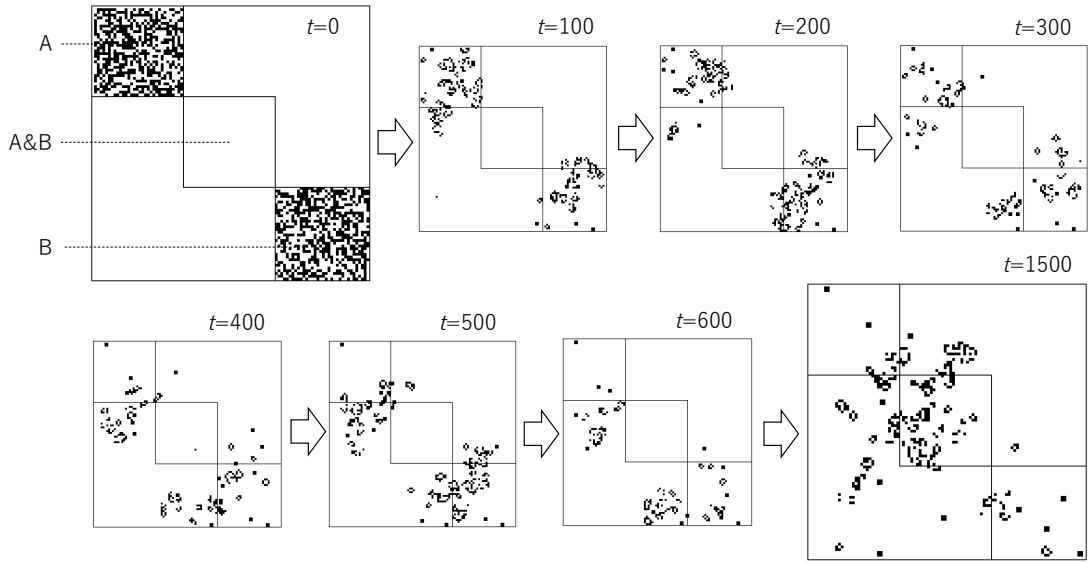

Figure 9. Typical time development of AND gate in asynchronous GL. The white arrow represents time development. The inner squares *A*, *B,* and *A&B* represent *S*(*A*), *S*(*B*), and *S*(*A&B*), respectively.

Figure 9 shows the computational process of the AND gate in asynchronous GL. Here, we define *T* = 1500, *N*=100, $P_1$=0.1 and $p_{NOISE}$ = 0.0001. This process implies that *A*=1 and *B*=1, thus *A&B*=1. The top left diagram represents an initial condition, where *A*=1 and *B*=1 are set with a density of 1 with a probability of 0.5, which exceeds $P_1$. After that, asynchronous GL is adapted to all cells, with the boundary conditions of which are that all cells have state 0. As time proceeds, broken gliders generated in *S*(*A*) and *S*(*B*) are propagated to the outside of *S*(*A*) and *S*(*B*), and finally (*t* = 1500 = *T*), the density of 1 for cells in *S*(*A&B*) exceeds $P_1$. This implies *A&B* = 1.

Figure 10 shows typical results of the AND gate in asynchronous GL. All parameters are set to be the same as those in Figure 9. Since the probabilistic logic gate shows the truth table with a specific probability, this result is just one typical example. Each pair of patterns connected by a blank arrow represents a pair of initial conditions and a final pattern at *t*=*T*. This implies that if *A*=0 and *B*=0, then *A&B* = 0 (top left in Figure 10); if *A*=0 and *B*=1, then *A&B* =0 (top right in Figure 10); if *A*=1 and *B*=0, then *A&B* = 0 (bottom left in Figure 10); and if *A*=1 and *B*=1, then *A&B* =1 (bottom right in Figure 10). Thus, the results in Figure 10 are consistent with the truth table of the Boolean AND gate.

The performance of this AND gate is discussed later.

While the implementation of the OR gate is the same as that of the AND gate, if the perturbation is different from the condition of the AND gate, the behaviors of the logic gate are totally changed. If $p_{NOISE}$ of the AND gate, 0.0001, is replaced by 0.001, the OR gate can be constructed. We abbreviate $A$ OR $B$ as $A||B$. Since broken gliders can be grown with a ten times larger perturbation ($p_{NOISE}$ = 0.001), a pair of input 1 and input 0 can generate much more broken gliders that can reach the output area, $S(A||B)$, than the situation with $p_{NOISE}$ = 0.0001. Thus, the probability of cells in $S(A||B)$ with state 1 can exceed $P_1$, which implies that the output is 1. In contrast, in the initial condition, $A = 0$ and $B = 0$, since there is no broken glider, but instead a cell with the state of 1 surrounded by cells with the state of 0, such isolated cells with the state of 1 cannot be grown to the broken gliders that can propagate to $S(A||B)$. That is why the result implies that output is 0.

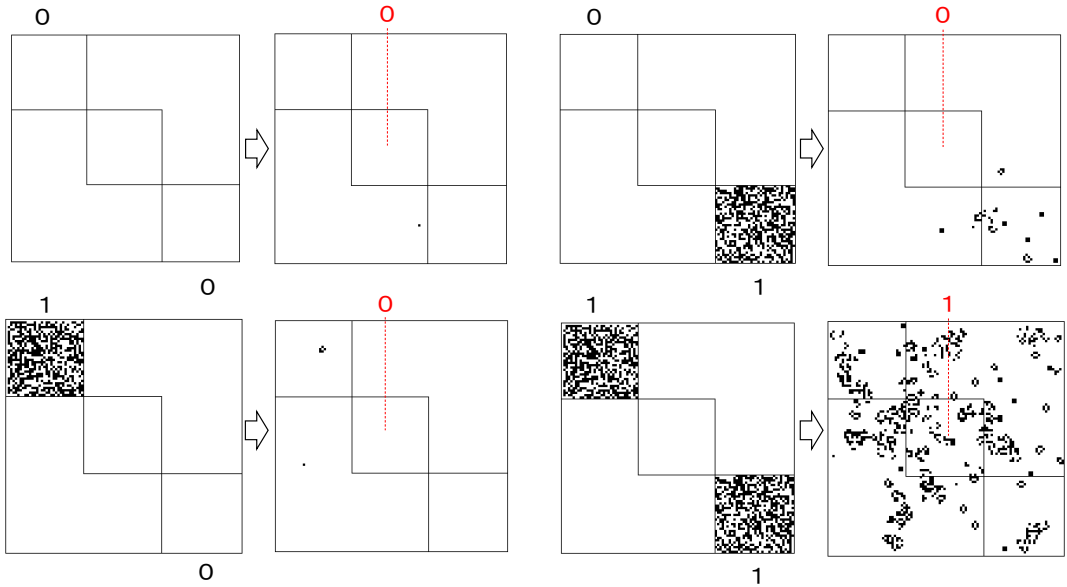

Figure 10. Computational results of the probabilistic AND gate implemented by asynchronous GL with perturbation, $p_{NOISE}$ = 0.0001, $P_1$=0.1 and $p$ = 0.13.

Figure 11 shows the computational results of the OR-gate implemented by asynchronous GL with the probability of stationary phase $p$=0.13. The input and output conditions are the same as those in the AND gate, although only the perturbation is different, such as $p_{NOISE}$ = 0.001. As shown in Figure 11, $A$=0 and $B$=0 lead to $A||B$=0; $A$=0 and $B$=1 lead to $A||B$=1; $A$=1 and $B$=0 lead to $A||B$=1; and $A$=1 and $B$=1 lead to $A||B$=1. Due to a large perturbation, broken gliders can reach $S(A||B)$. Cells with the state of 1 initially set as an input of 1 are first decreased and then are grown and propagated radially as if they had exploded. Thus, broken gliders are propagated anywhere, and the cells in state 1 could be grown in $S(A||B)$.

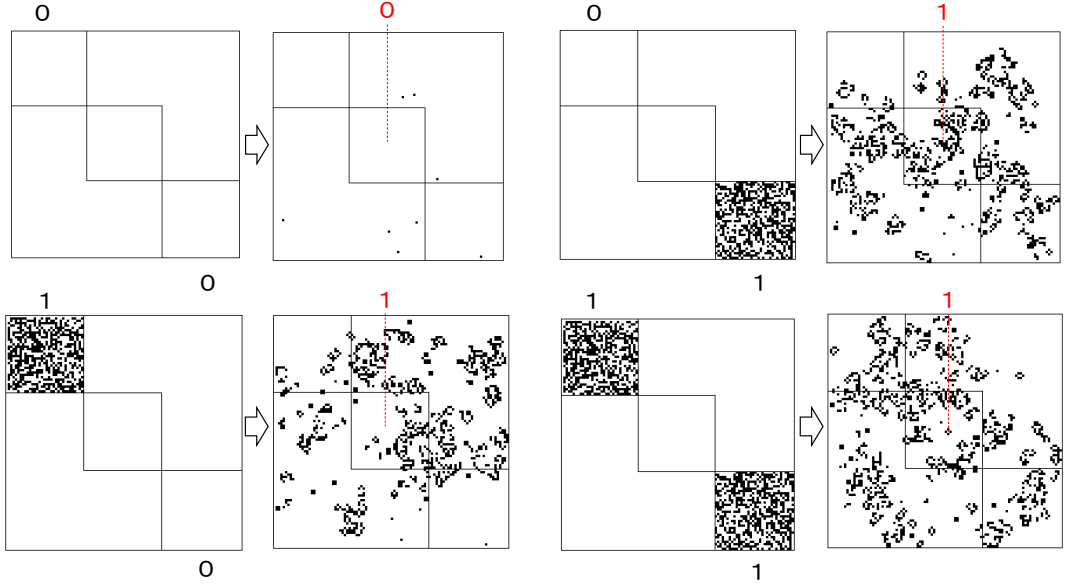

Figure 11. Computational results of the probabilistic OR gate implemented by asynchronous GL with perturbation, $p_{NOISE} = 0.001$, $P_1=0.1$ and $p = 0.13$.

Now, we evaluate the performance of the AND gate and OR gate implemented by asynchronous GL. The conditions of the two gates are the same as those of the gates in Figures 10 and 11. For each input condition of a pair of *A* and *B*, it is determined whether the number of cells with the state of 1 in $S(A\&B)$ or $S(A||B)$ exceeds $P_1$ (i.e., output =1) or not (output = 0) when $t = T$. The number of trials is 100 for each determinant, and the probability of output 1 is obtained as the number of instances of output 1 divided by the number of trials.

Figure 12 top left, four histograms show the frequency distribution of the cover of $S(A\&B)$. They show histograms for the conditions of *A*=0 and *B*=0, *A*=0 and *B*=1, *A*=1 and *B*=0, and *A*=1 and *B*=1. The horizontal axis represents the number of cells in $S(A\&B)$ whose state is 1 at *T* (i.e., the cover of $S(A\&B)$), where the actual number is obtained as the number multiplied by 20 (i.e., the cover of 3 represents 60). The vertical axis represents the frequency corresponding to the cover. The red line in the histogram represents $P_1$ multiplied by the number of cells in $S(A\&B)$, which implies that the cover on the right-hand side of the red line leads to an output of 1 and that on the left-hand side leads to an output of 0. Therefore, the summation of the frequencies on the right-hand side of the red line divided by the number of trials (100) is the probability of the output of 1. This results in the probability of the truth table corresponding to the AND gate, as shown in the top right of Figure 12. The probability of the truth table is expressed as a histogram whose horizontal axis represents a pair of inputs (00 represents *A*=0 and *B*=0; other abbreviations of numbers are similar), and the vertical axis represents the probability of the output of 1 corresponding to the input pair. The histogram shows that the probability of the output of 1 exceeds 0.5 only for the input of 11. This result is consistent with the truth table of the Boolean AND gate.

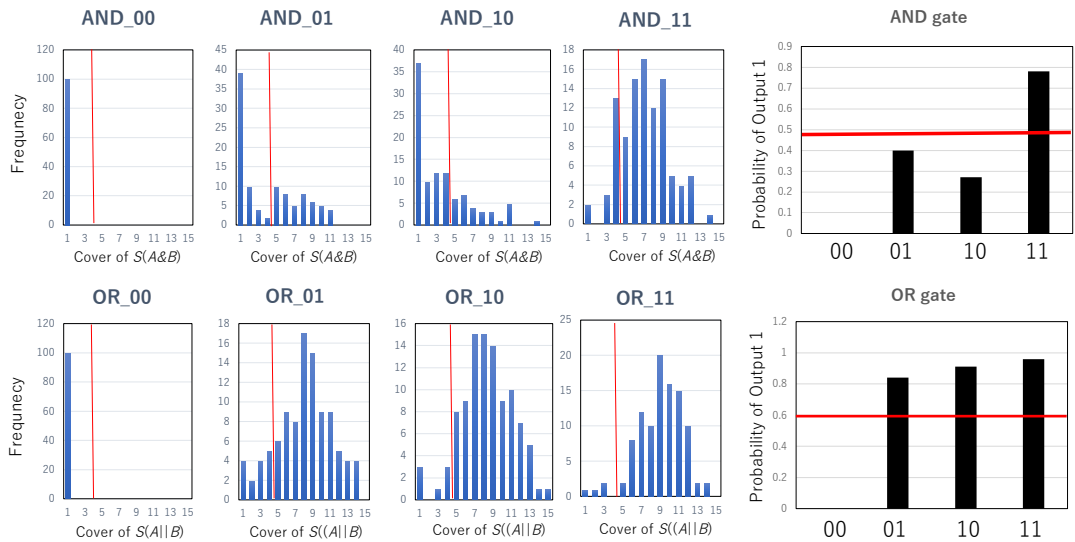

Figure 12. Frequency distribution of the cover of the output area (*S*(*A*&*B*) or *S*(*A*||*B*)) for each input pair (left four histograms) and the probability of the output of 1 against the input pair (right histogram). The top corresponds to the AND gate, and the bottom corresponds to the OR gate implemented by asynchronous GL.

The bottom left part of Figure 12 shows the frequency distribution of the cover of the output area, *S*(*A*||*B*), for inputs of 00, 01, 10, and 11 in the OR gate implemented by asynchronous GL. The red line is the same as that in the histogram for the AND gate. It is easy to see that most trials for inputs 01, 10, and 11 exceed the red line, indicating that $P_1$ is the threshold value for the output of 1. This leads to the histogram for the probability of the truth table corresponding to the OR gate, as shown in the bottom right of Figure 12. It shows that the probability of the truth table of the OR gate is consistent with that of the Boolean OR gate. Finally, we can conclude that both AND and OR gates can be implemented by asynchronous GL at the critical point.

3. Discussion

In the Introduction, we discuss the metaheuristic approach based on bioinspired computing, unconventional computing and self-organized criticality. It is known that those approaches can obtain quasi-optimal solutions for an ill-defined incomplete solution and that they frequently show critical properties characterized by the power law distribution. How can one manipulate a search method to obtain a quasi-optimal solution against perturbed conditions? This question is immediately related to the essential core of biological and/or unconventional computing.

Since there is no global clock in a biological system, interactions between components in a biological system are destined to be asynchronous. Although asynchronous computing is ubiquitous in natural computing systems, it has not been studied much, since its behaviors are different from synchronous computing, which has been extensively studied and is far from controllable. That is why implementing logic gates in asynchronous computing could be a touchstone to manipulating biological and/or unconventional computing.

In this paper, we take two-dimensional cellular automaton, game of life (GL), and study the behavior of GL in asynchronous updating. Although GL in synchronous updating can generate controllable gliders that can be used to implement logic gates, GL in asynchronous updating cannot generate controllable gliders. Although gliders generated in asynchronous GL are also unstable and rapidly collapse into fragments, they are regenerated due to the perturbation, and broken gliders can be propagated. Thus, those probabilistic gliders can be used to implement probabilistic logic gates.

First, we find that asynchronous GL shows a clear phase transition between the phase of oscillatory states and the phase of chaotic states. The parameter controlling phase is the probability of the stationary phase in asynchronous updating. In particular, we find that the critical point shows power law decay with an exponent of -0.1595. Since this exponent is found in various critical phenomena, it can be considered a universal property.

Second, we estimate the relationship between the computability and the criticality in asynchronous GL. Since logic gates in synchronous GL are constructed using gliders, the computability is evaluated with respect to the number of gliders generated. We find that most gliders can be generated in the region close to the critical point in the asynchronous GL space parameterized by the probability of the stationary phase. This suggests that critical phenomena have autonomous computability and/or self-organizing reachability to quasi-optimal solutions. It is well known that most biological systems show a power law distribution with various properties and that living biological systems are adapted to given environments. Based on those results, it can be said that systems showing critical properties have the ability to compute fitness in environments, which can make systems themselves able to adapt to the environments.

Finally, we construct the probabilistic logic gates implemented by asynchronous GL with a critical point. Since the generation and trajectories of gliders cannot be controlled in the logic gates, manipulating gliders is destined to be probabilistic. Manipulating logical values for logic gates is much more probabilistic than manipulating logical values in chemical computations such as BZ reaction computers. However, one can construct both AND gates and OR gates in asynchronous GLs.

One of the most intriguing aspects of probabilistic logic gates implemented by asynchronous GL is tuning noise in manipulating logic gates. While the essential mechanism and most of the conditions of the OR gate are the same as those of the AND gate, only the level of perturbation can make the difference between the AND gate and the OR gate. In other words, increasing perturbation can replace the AND gate with the OR gate. This result be an essential property of living systems, since the AND gate outputs the value 1 only for restricted input pair 1 and 1, while the OR gate outputs the value 1 for various input pairs, 1 and 0, 0 and 1, and 1 and 1. If the value 1 is interpreted as the active state of organisms, replacing the AND gate with the OR gate implies that replacing the active condition with the restricted conditions or the active condition with various conditions. In other words, it suggests that organisms utilize various resources if the environmental condition worsens (i.e., more perturbation). In this sense, the change from the AND gate to the OR gate by increasing perturbation is more biological than computational. If one can utilize the change from the AND gate to the OR gate corresponding to the change in the situation, a robust self-organizing computing could be implemented.

Since the performance of the logic gate is probabilistic, distributive laws such as $A\&(B||C) = (A\&B)||(A\&C)$ cannot always hold. It implies that the distributive law is weak. This could be related to the orthomodular lattice corresponding to quantum logic (Gunji et al., 2020a).

4. Conclusion

Derived from the idea of metaheuristic and/or self-organizing criticality, we investigate cellular automaton as a system carrying the essential core of biological and/or unconventional computing. Since the possibility of metaheuristics can be expressed as manipulating computations and the possibility of self-organizing criticality can be expressed as a power-law distribution, we investigate the behavior of an asynchronous game of life (GL) and estimate whether logic gates can be implemented in asynchronous GL.

We find that asynchronous GL shows a phase transition and that the GL at the critical point could have the highest computability with respect to the ability to generate gliders. Finally, we implement probabilistic AND gates and OR gates in asynchronous GL at the critical point, which can perform as logic gates to some extent. In our logic gates, perturbation plays an essential role in computing. Ten times multiplied perturbation makes the AND gate become the OR gate under the same conditions. Our study can be a touchstone to spell out the novel significance of biological and/or unconventional computing by showing the role of noise tuned to operate logic gates.

Acknowledgements: This work was financially supported by JSPS 18K18478, and JPJS 00120351748.

# References

Adamatzky, A 2010 Physarum Machines: Computers from Slime Mold. Word Scientific.

Adamatzky A (2018) On spiking behaviour of oyster fungi pleurotus djamor, Scientific reports 8 (2018) 1–7.

Adamatzky A, Tegelaar M, Wosten HA, Powell AL, Beasley AE, Mayne R (2020). On boolean gates in fungal colony, Biosystems 104138.

Aono M, Kasai S, Kim S-J, Wakabayashi M, Miwa H, Naruse M (2015) Amoeba-inspired nanoarchitectonic computing implemented using electrical Brownian ratchets. Nanotechnology 26, 234001. doi:10.1088/0957-4484/26/23/234001.

Bak P, Tang C, Wiesnfeld K (1987) Self-organized criticality: An explanation of 1/*f* noise. *Phys. Rev. Lett.* **1987**, *59*, 381–384.

Bak P, Tang C (1989) Earthquakes as a self-organized critical phenomenon. *J. Geol. Res. 94*, 15635–15637.

Bak P, Sneppen K (1993) Punctuated equilibrium and criticality in a simple model of evolution. *Phys. Rev. Lett. 71*, 4083–4086.

Bartumeus, F. Lévy processes in animal movement: an evolutionary hypothesis. Fractals 2007; 15: 151–162.

Berlekamp ER, Conway JH & Guy R. 1982 Winning Ways for Your Mathematical Plays, vol 2, Academic Press.

Chakrabarty S, Pal AK, Dey N, Das D, Acharjee S (2014) Foliage area computation using Monarch butterfly algorithm. In: Non-conventional energy (ICONCE), 2014 1st International conference on, 2014, pp 249–253.

Conrad M. 1983. Adaptability. New York: Plenum Publishing Corp.

Conrad, M (1985) On design principle for a molecular computer. Communication of the ACM 1985;

28(5): 464-80.

Cordero CG (2017) Parameter adaptation and criticality in particle swarm optimization. arXiv:1705.06966x1[cs.NE]19 May 2017.

Domany E, Kinzel W (1984) Equivalence of cellular automata to Ising models and direcgted percolation. Phys. Rev. Lett. 53, 311-314.

Dorigo M, Stützle T. Ant Colony, Ant Colony Optimization, MIT Press, 2004.

Erskine A, Hermann JM (2014) CriPS: Critical particle swarm optimization. Proc. Eur. Conf. Art. Life pp.207-14.

Fatès N (2001) Morvan, M.; An experimental study of robustness to asynchronism for elementary cellular automata. *Complex. Syst.*, *16*, 1–27.

Fatès N. (2006) Thierry, É.; Morvan, M.; Schabanel, N. Fully asynchronous behavior of double-quiescent elementary cellular automata. *Theor. Comp. Phys.*, *362*, 1–16.

Fatès N (2014). A guided tour of asynchronous cellular automata. *J. Cell. Autom.*, *9*, 387–416.

Fatès N, Morvan M (2005) An experimental study of robustness to asynchronism for elementary cellular automata. Complex Systems, 16, 1–27.

Gardner M (1970) The fantastic combinations of John Conways's New Solitaire Game 'Life'. Scientific American 223, 120-123.

Gunji, Y (1990). Pigment color patterns of molluscs as autonomy, generated by asynchronous automata. *Biosystem*, *23*, 317–334.

Gunji Y-P, Shirakawa T, Niizato T, Haruna T (2008) Minimal model of a cell connecting amoebic motion and adaptive transport networks. *J.Theor.Biol.*253, 659-667.

Gunji Y-P, Shirakawa T, Niizato T, Yamachiyo M, Tani I (2011) An adaptive and robust biological network based on the vacant-particle transportation model. *J. theor. Biol.* 272: 187-200.

Gunji Y-P, Nishiyama Y and Adamatzky A (2011) Robust soldier crab ball gate. *Complex Systems* 20, 94-104.

Gunji, YP (2014a). Self-organized criticality in asynchronously tuned elementary cellular automata. *Complex. Syst. 23*, 55–69.

Gunji, YP (2014b) Extended self-organized criticality in asynchronously tuned cellular automata. In: *Chaos, Information Processing and Paradoxical Games*; Vasileios, B., Ed.; World Scientific: Singapore.

Gunji YP, Shinohara S, HarunaT, Basios V. (2017) Inverse Bayesian inference as a key of consciousness featuring a macroscopic quantum logic structure. BioSystems 152, 44-63.

Gunji YP, Uragami D (2020) Breaking of the trade-off principle between computational universality and efficiency by asynchronous updating. *Entropy* 22(9) 1049.

Gunji, Y.P., Nakamura, K., Minoura, M., Adamatzky, A. (2020a). Three types of logical structure resulting from the trilemma of free will, determinism and locality. BioSystems, DOI information: 10.1016/j.biosystems.2020.104151.

Gunji, Y.P. Murakami M, Kawai T, Minoura M, Shinohara S. (2020b) Universal criticality beyond the trade-off in swarm models resulting from the data-hypothesis interaction in Bayesian inference. Computational and Structural Biotechnology (in review).

Hinrichsen H, Non-equilibrium critical phenomena and phase transitions into absorbing states. Advances in Physics 49(7), 815-958.

Humphries NE et al. Environmental context explains Lévy and Brownian movement patterns of marine predator. Nature 2010; 465: 1066-9.

Kauffman SA, Johnsen S (1991) Coevolution to the edge of chaos: Coupled fitness landscapes, poised states, and coevolutionary avalanches. J Theor Biol, 149(4); 467-505.

Nakagaki T, Yamada H, Toth A (2000) Maze-solving by an amoeboid organism. Nature 407, 470.

Nakagaki T, Iima M, Ueda T, Nishiura Y, Saigusa T, Tero A, Kobayashi R, Showalter K. (2007) Minimum-risk path finding by an adaptive amoebal network. Phys. Rev. Lett. 99, 068104. (doi:10.1103/PhysRevLett.99. 068104).

Nishiyama Y, Gunji Y-P, Adamatzky A (2012) Collision-based computing implemented by soldier swarms. *Int. J. Parallel, Emergent and Distributed Systems*. **DOI:** 10.1080/17445760.2012.662682.

Nordfalk J & Alstrøm P (1996) Phase transitions near the "game of life", Phys. Rev. E 54(2), R1025-1028.

Rennard J-P (2002) Implementation of logical functions in the Game of Life. In: Collision-Based Computing (Adamatzky A ed.), pp. 491-512.

Reynolds AM, Schultheiss P, Cheng K (2013) Are Lévy flight patterns derived from the Weber–Fechner law in distance estimation? Behav Ecol Sociobiol, DOI 10.1007/s00265-013-1549-y

Reynolds AM (2018) Current status and future directions of Lévy walk research, Biology Open 2018; 7: bio030106.

Sims DW et al. Scaling law of marine predator search behaviour. Nature 2008; 451: 1098-1102.

Takamatsu A, Tanaka R, Yamada H, Nakagaki T, Fujii T, Endo I. (2001) Spatiotemporal symmetry in rings of coupled biological oscillators of Physarum plasmoidal slime mold. Phys. Rev. Lett. 99, 068104.

Tero A, Takagi S, Saigusa T, Ito K, Bebber DP, Fricker MD, Yumiki K, Kobayashi R, Nakagaki T. (2010) Rules for biologically inspired adaptive network design. Science 327, 439 – 442. (doi:10.1126/science.1177894).

Toth R, Stone C, Adamatzky A, Costello BL, Bull L (2009) Experimental validation of binary collisions between wave fragments in the photosensitive Belousov-Zhabotinsky reaction. Chaos, Solitons & Fractals 14(4), 1605-1615.

Tsuda S, Aono M, Gunji Y-P (2004) Robust and emergent *Physarum*-computing. *BioSystems* 73, 45-55.

Viswanathan GM, Suldyrev SV, Havlin S, de Luz MGE, Raposo EP, Eugene Stanley H (1999) Optimizing the success of random searches. Nature, 401: 911–4.

Wang G-G, Guo L, Gandomi AH, Hao G-S, Wang H. (2014) Chaotic Krill Herd algorithm. Information Science 274, 17-34.

Wang G-G, Deb S, Coelho LDS (2018) Earthworm optimization algorithm: a bio-inspired metaheuristic algorithm for global optimization problems. International journal of bio-inspired computation 12(1):1, DOI:10.1504/IJBIC.2018.093328.

Zhu L, Kim S-J, Hara M, Aono M. (2018) Remarkable problem-solving ability of unicellular amoeboid organism and its mechanism. R. Soc. open sci. 5: 180396. http:// dx.doi.org/ 10.1098/ rsos.180396.